\pdfoutput=1
\documentclass{svproc}
\usepackage{url}
\usepackage{subcaption}
\usepackage{graphicx}

\begin{document}
\mainmatter              
\title{Lightweight neural networks for affordance segmentation: enhancement of the decoder module}
\titlerunning{Lightweight neural networks for affordance segmentation}  
%
\author{Simone Lugani\inst{1} \and Edoardo Ragusa\inst{1} \and 
Rodolfo Zunino\inst{1} \and Paolo Gastaldo}
\authorrunning{Edoardo Ragusa et al.} 
%
%
\institute{Department of Electrical, Electronic, Telecommunications Engineering and Naval Architecture (DITEN),
University of Genoa, Genoa, Italy\\
\email{edoardo.ragusa@unige.it},\\ 
}

\maketitle              

\begin{abstract}
The deployment of deep neural networks for visual affordance segmentation on wearable robots poses may prove critical, due to some conflicting aspects of the problem. On one hand, affordance segmentation requires high-level abstraction capabilities, that typically involve large-size models. On the other hand, computing resources hosted on wearable robots prevent to run large-size models in real-time. The paper presents an analysis of the role of the segmentation head in the trade-off between generalization performance and compute cost. The obtained models outperform modern baseline solutions in well-known, real-world datasets while meeting low computing requirements.

\end{abstract}
\section{Introduction}

Wearable robots such as prostheses and exoskeletons are advanced systems with multiple degrees of freedom. The control of these devices is an open problem because the operation of all functions requires to generate generating explicit commands \cite{salminger2022current}. 

Semiautonomous control automatizes parts of the action by reducing the users’ efforts  \cite{tang2022wearable,sun2020real}. Toward that end, the system selects the action autonomously; this implicitly requires high-level capabilities in both semantic reasoning and fine-grained sensing. Teleceptive sensing \cite{krausz2019survey}, i.e. sensing without contact, enables sensing information from the environment, for example by using cameras. Visual affordance segmentation (VAS) divides an object into functional parts, thus paving the way to planning fine-grained semiautonomous actions. State-of-the-art approaches use deep neural networks (DNNs) \cite{nguyen2017object,jiang2021synergies,khalifa2022towards} to tackle this problem, but the typical hardware resources embedded in wearable robots prevent real-time processing. As a result, the design of those solutions combines very complex computer vision problems with tight computing requirements. In \cite{ragusa2021hardware} the authors proposed a simplified version of the original VAS problem supported by portable hardware. The original solution was subsequently improved by proposing additional DNNs modules \cite{apicella2021affordance} and sensory information \cite{ragusa2023affordance}. 

Balancing generalization performance and hardware requirements is an open problem, further complicated by many levels of abstraction. When designing DNNs, most approaches take into account the design of the backbone part of the architecture, for example by using Network Architecture Search (NAS) \cite{benmeziane2021comprehensive}. This choice seems reasonable when considering that the backbone plays a crucial role when compared with the segmentation head (SH), in terms of both computations and generalization performance. However, the SH plays a significant role that has been under-explored in the specific scenario considered in the research presented here.

This paper considers the impact of the Segmentation Head components on visual affordance segmentation when using hardware-friendly DNNs. The analysis first shows that even minor adjustments in the SH architecture can affect the overall performance of the model. Secondly, a hierarchical problem formulation based on multitask learning leads to improved generalization performance. The resulting set of hardware-efficient SHs outperform existing solutions on established benchmarks in terms of accuracy, while ensuring a comparable computational cost. 

\section{Method}
Small-size networks may typically exhibit reduced generalization
ability as compared with large-size models \cite{ragusa2021hardware}. As a consequence, the role of the SH becomes critical, as the backbone might fail to disentangle effectively the implicit features involved in the problem. 

Segmentation Heads accomplish two basic tasks: 1) the reconstruction or rearrangement of geometrical information, based on the features extracted by the backbone levels; 2)  the eventual classification of affordable parts. The literature proposes two main solutions for the reconstruction of geometrical information. Standard upsampling operations, such as bilinear upsampling, can augment image resolution, and are not involved in the solution of the learning problem. Conversely, transposed convolution layers perform a convolution operation with a stride value smaller than one, thus yielding larger output tensors. In this setup, the kernel plays a crucial role in the learning process, at the expense of a larger number of flops with respect to standard upsampling. A few convolutional layers typically support the eventual classification step. Standard convolution exhibits satisfactory expressive capabilities, whereas depth-wise separable schemes can better balance computing requirements and generalization capabilities. 



Object localization is critical in VAS \cite{nguyen2016detecting}, as it is preliminary to understanding the object's affordable parts. One can formalize a hierarchical version of the learning problem, where the network includes two output heads. The first SH supports a binary Object segmentation (OS) classification task. The computational graph includes an additional SH, which tackles the actual VAS problem. This component processes an input representation that is optimized to isolate the object, and focuses on the distinctive parts of the object, thus biasing the feature extraction process. The loss function is formalized as:
\begin{equation}
    \mathcal{L} = \mathcal{L}_{seg} + \mathcal{L}_{aff}
\end{equation}
where the two terms refer to OS and VAS, respectively. 

\subsection{Segmentation head}
\begin{figure}
    \centering
\includegraphics[width=0.5\textwidth]{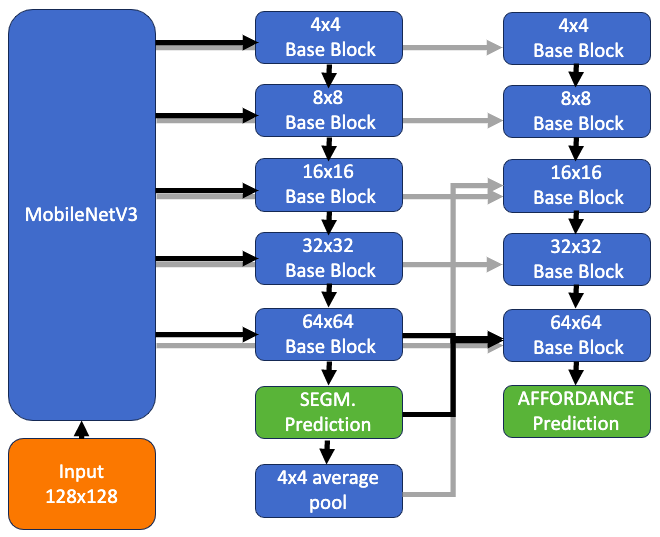}
    \caption{General scheme of the proposed SHs.}
    \label{fig:temp}
\end{figure}

The large networks, which yet fit the memory constraints of embedded devices, typically feature inference times of many seconds \cite{canepa2022detection}, which proves inadequate for AS. FLOPs have a major impact on inference time and can be computed independently from the target platform \cite{benmeziane2021comprehensive}. Therefore, the FLOPs count was selected as the constraint in the SH design process. When paired with the MobileNetV3 backbone for one image of size 128x128, the admissible SHs were limited to the ones with a FLOP count in the range of 700/800 M. Figure \ref{fig:temp} shows the template architecture for a multitask scenario. In the case of a single task, one just removes the OS branch. The subset of connections with the backbone is selected based on the solution selected for the \textit{base blocks} to control the number of FLOPs. The base blocks always contain a convolutional and an upsampling layer. 

The research presented in this paper compared three instances of the base block, namely, depthwise separable convolution with nearest upsampling (U),  depthwise separable convolution with transposed convolution layer (T), and standard convolution and nearest upsampling (B). 

\section{Experiments}
The experiments were divided into three sets. The first experimental section compared the segmentation heads; the second set of tests measured the accuracy values scored for the auxiliary task. Finally, the role of the interconnection between SH and backbone was considered.

Two datasets provided the experimental benchmarks: the UMD dataset held 28.843 RGB-D images of 7 categories of objects. 5135 images formed the test set. Likewise, the IIT dataset (IIT) \cite{nguyen2017object}, including 8835 images at different framing and resolution settings, was used for the assessment.  In compliance with the setup described in \cite{ragusa2021hardware}, all the grasping affordances were grouped into a unique class and the other affordances were grouped into a single category (i.e. "don't grasp").  

All models were trained on both training sets. The IIT dataset was augmented by applying standard geometrical and color distortions. The UMD was augmented by a custom procedure, replacing the original blue background in the images with a different picture. The objects were isolated by using the segmentation mask of the labels and a color based-procedure relying on the uniform background. Finally, standard augmentation was applied. 

The baseline SH relied on standard convolution and nearest upsampling, and adopted the design strategy presented in \cite{ragusa2021hardware}. In the following, the architectures with multiple SHs will be denoted by a letter for each head. For example, decoder 'UB' will denote an architecture including Segmentation Head 'U' and Visual Affordance Segmentation 'B'. All the SHs were connected to the backbone using four connections [with reference to figure \ref{fig:temp}, the 64x64 connection was not implemented]. 

Class-wise weighted pixel-wise accuracy was computed over the test set. Three test sets were considered: the IIT test set, the original version of UMD, and the version with the replaced background UMD\_B. A summary performance measure will be indicated with TOT computed as the weighted average of the three accuracies, weighting IIT 0.5 and the two versions of UMD 0.25. 

Table \ref{tab:ASper} gives the weighted accuracy of the networks on the three versions of the test sets, also including the number of parameters. All the tested SH outperformed the baseline approach. The baseline implementation relied on a different subset of connections and prioritized high-level features, thus leading to an increased number of parameters. The networks with 'B' heads benefited from the use of standard convolution, leading to the reported good results. The two architectures based on the single head with depthwise convolution both obtained excellent scores; this proved that, even without multitasking, depthwise segmentation could support the segmentation process effectively.

\begin{table}[]
    \centering
    \begin{tabular}{lcccc}
    \hline
       & TOT& IIT & UMD & PARAMS. \\[2pt]
\hline\rule{0pt}{12pt}
    Decoder B   & 91.8  & 88.6 & 94.6 & 2.2 M \\
    Decoder U   & 92.6  & \textbf{91.0} & 94.6 & \textbf{1.3 M} \\ 
    Decoder T   & 92.6  & 90.5 & 95.1 & 2.0 M \\
    \hline
    Decoder BB  & 92.5  & 89.9 & \textbf{95.6} & 3.6 M \\
    Decoder UU  & 92.1  & 89.6 & 94.9 & 1.6 M \\
    Decoder TT  & 92.5  & 90.3 & 95.2 & 2.1 M \\
    Decoder TB  & \textbf{92.8}  & 90.8 & 95.2 & 3.6 M\\
    Decoder UB  & 91.7  & 90.3 & 95.2 & 3.3 M\\
    \hline
        Baseline \cite{ragusa2021hardware}   & 91.2  & 88.6 & 94.6 & 5.9 M \\
    \hline
    \end{tabular}
    \caption{Affordance segmentation performance}
    \label{tab:ASper}
\end{table}

Table \ref{tab:OSper} presents the results of multitask architecture for the object segmentation task. In this case, the result on the augmented UMD\_B dataset has been added. The Table uses the same convention of Table \ref{tab:ASper}. The results confirmed that all the SHs managed to discriminate the object pixels from the background pixels with an accuracy higher than 90\% on average. 

\begin{table}[]
    \centering
    \begin{tabular}{lccccc}
       & TOT & IIT & UMD & UMD\_B \\[2pt]
\hline\rule{0pt}{12pt}	
    Decoder BB  & 92.7 & 90.2 & 95.4 & 95.0  \\
    Decoder UU  & 93.1 & 90.2 & 96.0 & 95.9 \\
    Decoder TT  & 93.7 & 91.3 & \textbf{96.2} & \textbf{96.1} \\
    Decoder TB  & \textbf{93.9} & \textbf{91.8} & \textbf{96.2} & 96.0 \\
    \hline
    \end{tabular}
    \caption{Object segmentation performance}
    \label{tab:OSper}
\end{table}

Table \ref{tab:connections} illustrates the results of the experiments about the connections between the SH and the backbone, comparing a new version of the SHs that used the 64x64 layer as a connection. The 8x8 connection was removed in this setup while maintaining four levels of feature maps. Table \ref{tab:connections} follows the conventions adopted in  Table \ref{tab:ASper} to report the performance of the new versions of the SHs; the differences with respect to the previous versions are reported in brackets. The results confirmed that exploiting connection with low-level features did offer some improvements, although the previous feature set actually yielded the best overall average result. This result confirmed the importance of connections, and highlighted that they should represent a crucial factor in automated design strategies. 

\begin{table}[]
    \centering
    \begin{tabular}{lcccccc}
        \hline
                    &   TOT   & IIT & UMD & PARAMS \\    [2pt]
\hline\rule{0pt}{12pt}	
       Decoder UT-X &   \textbf{92.7} (+1.5)     & 90.5 (+2.4) & \textbf{95.4} (+0.7)      & \textbf{1.1M} (-0.6M) \\
       Decoder TU-X &   92.6 (+0.5)     & \textbf{90.7} (+1.5) & 94.9 (-0.6)      & 1.4M (-0.6M) \\
       Decoder TT-X &   92.6 (+0.1)     & 90.5 (+0.2) & 95.1 (-0.1)      & 1.5M (-0.6M) \\
       Decoder TB-X &   92.6 (-0.2)     & 90.5 (-0.3) & 95.3 (+0.1)      & 1.8M (-1.8M) \\
       Decoder BB-X &   92.5 (+0.0)     & 90.3 (+0.4) & 95.1 (-0.5)      & 1.8M (-1.8M) \\
           \hline
       \end{tabular}
    \caption{VAS performance with low-level feature maps}
    \label{tab:connections}
\end{table}

\section{Conclusion}
The paper presented an analysis of the role of the segmentation head for lightweight neural networks in applications of visual affordance segmentation. The paper analyzed the problems that may arise when using constrained networks, and presented some design criteria to limit the issues introduced by the reduced generalization capabilities of the models. Empirical results confirmed that a careful design of the segmentation head, with a slight reformulation of the learning problem, could improve over the baseline solutions proposed in recent works.    
%
\bibliographystyle{splncs}
\bibliography{references}
\end{document}